# An Image captioning algorithm based on the Hybrid Deep Learning Technique (CNN+GRU)

Rana Adnan Ahmad
Department of Computer Science
Comsats university islamabad, sahiwal campus,Sahiwal,Pakistan
rana22293@gmail.com

Muhammad Azhar
Department of computer science
Comsats university islamabad, sahiwal campus,Sahiwal,Pakistan
azhar.cs@cuisahiwal.edu.pk
Chosun University, Republic of Korea
Muhammad.azhar@chosun.ac.kr

Hina Sattar
Department of computer science
Comsats university islamabad, sahiwal campus, Sahiwal, Pakistann
hinasattar987@gmail.com

*Abstract*— Image captioning by the encoder-decoder framework has shown tremendous advancement in the last decade where CNN is mainly used as encoder and LSTM is used as a decoder. Despite such an impressive achievement in terms of accuracy in simple images, it lacks in terms of time complexity and space complexity efficiency. In addition to this, in case of complex images with a lot of information and objects, the performance of this CNN-LSTM pair downgraded exponentially due to the lack of semantic understanding of the scenes presented in the images. Thus, to take these issues into consideration, we present CNN-GRU encoder decode framework for caption-to-image reconstructor to handle the semantic context into consideration as well as the time complexity. By taking the hidden states of the decoder into consideration, the input image and its similar semantic representations is reconstructed and reconstruction scores from a semantic reconstructor are used in conjunction with likelihood during model training to assess the quality of the generated caption. As a result, the decoder receives improved semantic information, enhancing the caption production process. During model testing, combining the reconstruction score and the log-likelihood is also feasible to choose the most appropriate caption. The suggested model outperforms the state-of-the-art LSTM-A5 model for picture captioning in terms of time complexity and accuracy.

*Index Terms— Deep Learning, Image captioning, CNN, GRU*

## 1. INTRODUCTION

Deep Learning has made great strides recently due to rapid growth and high utilization [1-4]. Thus, similar to Neural Machine Translation (NMT)[5], generating captions of the images through neural encoder-decoder framework has shown the dominance in recent years. In this process of image captioning, encoding of the image is done through encoder which is typically from the Convolutional Neural Networks (CNN) [6] family (like Vanilla CNN [6], Region based CNN [7], Fast R-CNN [8], Faster R-CNN [9] etc.) and decoder is from the RNN family [10] (like LSTM [11], BLSTM [12] etc.). In this framework of CNN-LSTM pair [13-16], the encoder (CNN) learns the visual features by making the feature maps and max-pooling during the feature learning stage and then detection of objects after flattening and applying fully connected layer. Thus it converts the image to vector of numbers which is learned form of the visual content of the image under consideration. In the decoder part, the vector output of the encoder is used as the initial input of the decoder to produce caption word by word. Even though Long Short Term Memory (LSTM) solves the issue of handling long dependency by decreasing the effect of exploding and vanishing gradients [11], the time complexity issue is still a major drawback in this model due to many gates residing in the LSTM unit for the memorization purpose. Another key issue with these kind of encoder-decoder models are the lack of understanding of the semantic context as the encoder of these models fail to transfer the major key visual information to the decoder. Because of the absence of reverse dependency checking (Caption-to-Image), these models do not perform well in case of complex images.

Several approaches have been proposed to deal with the above-mentioned issues [17-20]. Some researches have proposed the attention mechanism to get the information from the key regions automatically and tried to encode that specific information into the context vector which then used by the decoder to generate the caption [17,18]. Some other researchers have tried to extract semantic attributes as a supplement of the CNN features to embed into encoder by various methods [19, 20].

The major drawback of all the above mentioned methods was that those methods only explore the image-to-caption dependency but not the reverse way for the validation of the extracted information. Even though, Jinsong Su et.el. [21] have tried to use the semantic reconstructor of caption-to-image but still they could not validated the results in effective way. In addition to this, the time complexity issue was also remained due to the usage of LSTM unit.

To resolve the above mentioned issues, we have proposed a hybrid deep learning technique based on the CNN-GRU encoder-decoder gramework with the better hyper-parameter tuning and with the caption-to-image validation method by taking the motivation from Jinsong Su et.el. [21]. This caption-to-image reconstructor helps to handle the semantic context into consideration as well as the time complexity. By taking the hidden states of the decoder into consideration, the input image and its similar semantic representations is reconstructed and reconstruction scores from a semantic reconstructor are used in conjunction with likelihood during model training to assess the quality of the generated caption.

As a result, the decoder receives improved semantic information, enhancing the caption production process. During model testing, combining the reconstruction score and the log-likelihood is also feasible to choose the most appropriate caption.

To validate our proposed method, we have used the benchmark MS COCO dataset [22] and the experimental results have proved that our method outperformed the current state-of-the-art methods in terms of accuracy and time complexity.

## 2. RELATED WORK

Inspiration for our work comes from the auto encoder [23,24] and how well it performs in NMT [25], which employs semantic production to hone the learning representation of input data. In this activity, we are fine-tuning the idea with captions to the image. Basically, related work involves taking after two strands. In general, NMT's common hands are very much based on the demonstration of source-to-target interpretation. Encouraged by questions about the auto encoder that makes reproduction more realistic and looking at whether the recreated inputs are more reliable than the original inputs [26], many analysts are committed to using the adaptation of dual-directed NMT conditions [27].

Compared with NMT, most models of neural image captions are based on the neural encoder-decoder system [30]. However, this engineer cannot guarantee that the image data can be completely converted into a decoder. To discuss this problem, analysts are currently accepting to take after two types of approaches: (1) As in NMT attention [31], a few analysts link part of the visual considerations to capture the semantic presentations of critical image regions [32,33]. (2) In various ways, a number of analysts are committed to extract semantic features or high-level concepts into images, which can be integrated into an LSTM-based decoder as an additional input [28,29]. In this way, the show will be directed to settings that are closely related to the theme of the image. Besides, You et al. [34] encompassed the two types of methods listed above.

Our proposed representation is based on the CNN-LSTM model, in which the proposed semantic reconstructor is comparably compared to the LSTM, which is why it benefits both to display preparation and testing when the regional language indicate and the coding system are modeled independently. From Wena et al. [35] Institution devoted to improving automatically generated image captions by making inferences about their semantic content. However, the visual highlights are generally employed as the decode of the decoder in the current model captions, while the semantic elements of the image are provided exclusively to the decoder. As a result, we agree that visual robustness is more crucial than semantic characteristics. Through this research, we provide semantic features to neural machine translation as well as video captions. In conclusion, we are experimenting with three different methods for reconstructing photos based on fabricated captions. Additionally, our claim may be distinct from earlier studies due to K's extensive utilization of visually similar photos.

3. PROPOSED MODEL

This section describes the proposed hybrid deep learning approach based on CNN-GRU encoder-decoder framework. This framework has 3 major parts, 1) Encoder: which is the CNN. 2) Decoder: which is the GRU layer and 3) The Semantic validator: for validation of the caption-to-image information.

*A. Model architecture*

The three neural network modules (Encoder, Decoder, and Semantic validator) that make up our proposed model are depicted in Fig. 1. The details of each module is given below:

- Encoder

In encoder, a model similar to [36] has been used where the image *I* is taken as input and the features from the image **F** is extracted by the CNN-based encoder. The feature vector **F** $\in R^{D_v}$ is used to represent the features extracted from the image *I*. $D_v$ represents the diemnsions of the feature vector. As all the sementic information can not be extracted by one feature vector, thus additional semantic attributes have been extracted by the algorithm proposed by Yao et al. [36]. The extracted attribute vector is denoted by **A** $\in R^{D_a}$ which shows the probabilty of each high level attribute existed in the caption dataset which is generated by the MIL (Multiple Instance Learning) model presented in [27]. MIL model showed the promising results in finding the semantic relations between the attributes of the image. Da represents the diemnsions of the attribute vector **A**.

After extracting both feature map **F** and attribute vector **A**, the encoder gives these 2 outputs to decoder as an input which is used for the caption generation purpose.

- Decoder

As we got feature vector **F** and the attribute vector **A** as an output of the encoder from previous network, this F and A is used as the input to the decoder for the caption generation. Yao et al. [36] proposes 5 different and diverse variants for the LSTM network and it is proved that the fifth one named LSTM-A$_5$ works better than others, so we also used the same network for getting the better performance. Thus, according to LSTM-A$_5$, we used the **A** and **F** vectors to calculate the log Probability $\mathcal{E}$ as mentioned in Equation (1).

$$\mathcal{E}(S|I) = \mathcal{E}(S|F, A) = \sum_{t=1}^{N_s} \mathcal{E}(w_t | F, A, w_{<t}) \quad (1)$$

Where **F** and **A** represents the feature vector and attribute vector respectively. **S** is the set of words generated by the attribute vector **F**. **S**= {$w_1, w_2, \ldots w_{N_s}$} and Ns is the size of the set S. **I** is the actual image.

The log probability $\mathcal{E}(w_t | F, A, w_{<t})$ is directly proportional to the expection of ($w_t^T$ E ($vh_t$ + b) as shown in Equation (2).

$$\mathcal{E}(w_t | F, A, w_{<t}) \propto \exp(w_t^T \ E \ (vh_t + b)) \quad (2)$$

Where **E** represents the matrix of the word embeddings, v denotes the corresponding matrix while b shows the bais. $h_t$ is the hidden state. The hidden state calculation is discussed in detail in [7].

- Semantic validator of caption to image

As shown in Fig. 1, the semantic redesign of the description to picture work to recreate the semantic demonstration of every single input image since its comparative captions.

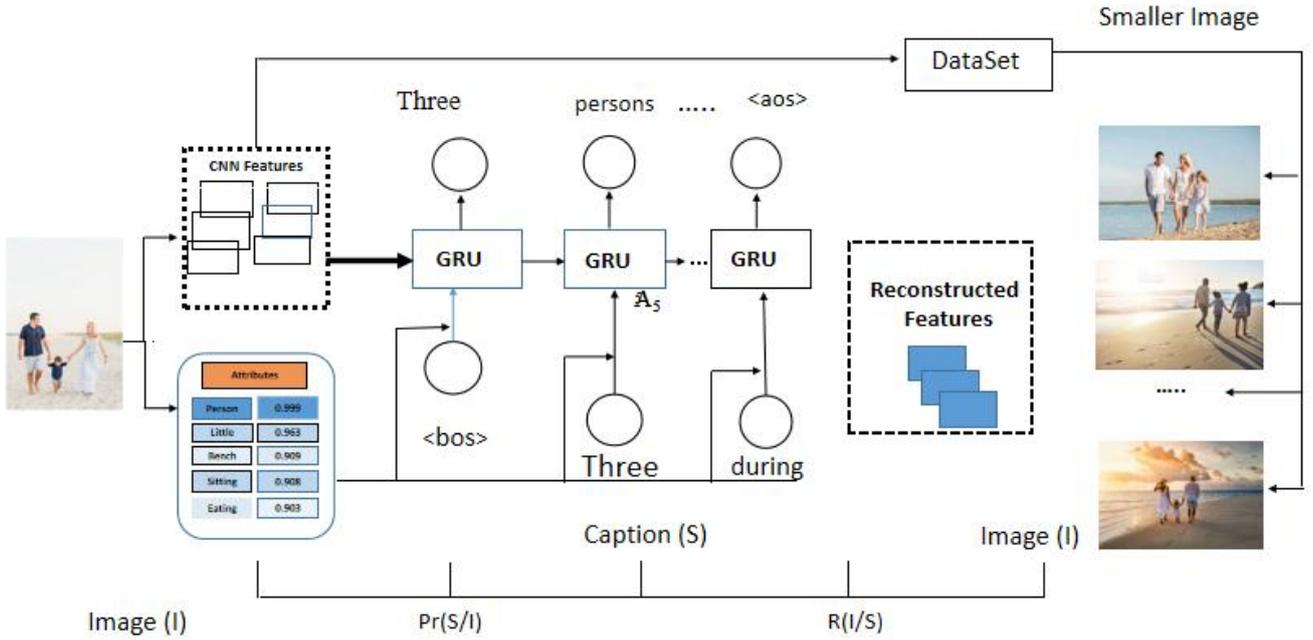

Figure 1: Provides an overview of our proposed model's architecture, which consists of three neural networks (Encoder, Decoder, and Semantic Reconstructor).

Naturally, a complete semantic reconstructor must meet the subsequent dual requirements: On the other side, its reconstructed presentation must be precise and sufficient to replicate image data; on the other side, its use is not compiled to have the greatest impact on professionalism. In this case, we are referring directly to the caption S that has the coverings h = {h1, h2,. . ., hNs} play a significant part in the description era. At that point, in this framework, we are examining three semantic functions to determine the semantic demonstration of the created captions, represented by hc, which can help to recreate the reconstituted direction of the input image, labeled Ir.

- Model Training

The training set $D_{train}$ = {(F, A, S)}, is used to generate the objective function as follows:

$$O(D; \theta_{ed}, \theta_{dr}) = \arg\max(\theta_{ed}, \theta_{dr}) \sum_{(I,S)\in D} \{\mathcal{E}(S|I; \theta_{ed}) + \lambda \cdot R(\{I1, \ldots, IK\}|S)\} \quad (3)$$

So we have to maximize the reconstruction score based on $\theta_{ed}, \theta_{dr}$. $\theta_{ed}, \theta_{dr}$ represents the encoder and decoder model parameters. λ is the hyper-parameters.

- Model Testing

During testing, semantic reconstruction can be utilized to improve selected captions. As is it shown in Fig 2, we use a multi-stage system that combines beam search and position reset. Inserted image captioning techniques:

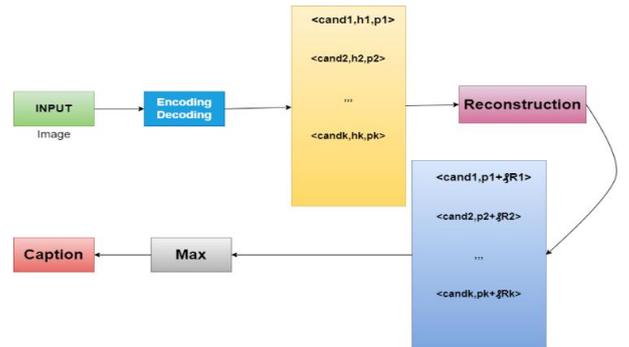

Figure 2: A test image of our model. h, P, and R show the hidden sequence, log likelihood probability, and caption reconstruction score, respectively.

1. A collection of applicant captions, log possibilities, and unobserved state sequences are generated using the standard decoder components via an initial application beam investigation.
2. After that, we use the hidden captions of each candidate to reconstruct the semantic model of the merged image by computing the appropriate reconstructive points.

3. After arranging a log and potential school rebuilding sites, we calculate the final outcome of each caption and select final captions based on the combination of points.

## 4. EXPERIMENTS

The experiments have been conducted on the most popular benchmark dataset COCO [30] to compare the performance of our image captioning proposed model with other state-of-the-art methods.

- Experimental setup

COCO data-set was used to check the validity of our proposed model which contains 130000 manually annotated images. Each image has 4 descriptions which were used for the training purpose. In addition to this, 5000 images were used as testing dataset.

Out of the 130000 images of the training set, 80000 images were used for the training purpose while 5000 images were used for the validation purpose. Based on these settings, the vocabolary has been built with 8500 unique words. For getting the image features, the following setting was used for the hyper-paramter tuning.

Adam [32] is used as the optimizer. We employed stop-reading techniques [33] and pre-stop techniques, and we determined the following take-out hyper-parameter parameters: reading level beginning at 2 4, input rate as 300, covering layer size as -1024, mini-batch. A maximum cycle count of 30 is used with a scale of 1024. We used Word2vec's [34] pre-trained embeddings, which we optimized by setting the tradeoff parameter to 1. The threshold was established at 3 in our model testing.

- Evaluation metrices used:

The evaluation metrices used are 1) BLEU [40] where we set beam size K=3 thus BLEU@1, BLEU@2, BLEU@3 and BLEU@4 are calculated. In addition to this, METEOR [46] which is shown as M in Table 1, ROUGE-L [37] which is shown as R and CIDEr-D [38] which is shown as C in the Table1. The values of these metrices were calculated by the COCO released code [39]. BLEU, ROUGE-L, and METEOR were initially developed as benchmarks for evaluating the accuracy of machine translation. Image caption testing follows the same procedure as machine translation testing, where the phrases generated are compared to the actual sentences, and metrics are often utilized.

- Description of the compared state-of-the-art methods

1) NIC: The decoder of NIC is based on LSTM which directly use features of the images as input to LSTM.
2) ME: The distinction of this method is its language model that explore the mappings bidirectionally in images and their captions. This language model is independently built from the encoder-decoder framework.
3) ATT: This model uniquely extracts the key information of the images by a model based on semantic attention.
4) Soft-Attention and Hard Attention (SA and HA) models: This model differs from other models in terms of using CNN features as input to decoder. The Soft-Attention (SA) is with the normal Back-propagation method while in Hard-Attention (HA), the stochastic attention is used with re-inforcement learning.

5) LRCN: It is unique in terms of taking the image feature and its previous caption as the input at each time-step.
6) Sentence Condition (SE): In this method, a text-conditional attention model is used which helps decoder to learn the semantic information of the text.
7) LSTM-A5: This is based on the best variant of LSTM. Our propsoed model is inspired by this. We have used the same settings of LSTM-A5 for comparison purpose as the dataset is also same.

- Test results on COCO

The results got from the experiments by using the COCO dataset is shown in the Table 1. As it is obvious from the results, our method performed better than all other state-of-the-art methods. The results of the metreces BLEU@1, BLEU@2, BLEU@3 and BLEU@4 are all better than the NIC, HA, SA, ATT, ME, and other compared methods. Even on the metrics METEOR [37] which is shown as M in Table 1, ROUGE-L [38] which is shown as R and CIDEr-D which is shown as C in the Table1, which were initially developed as benchmarks for evaluating the accuracy of machine translation, our resulting indexes are still better on the above metrics as compared to NIC, HA, SA, ATT, ME, and other compared methods.

These results proved that the proposed CNN-GRU method with semantic validator of caption-to-image is working perfectly.

Table 1: The performance of our proposed model against other state-of-the-art methods building VGG framework or GoogleNet framework. For clarity, B@K is for BLEU@K where K={1,2,3,4}, MET is used for METEOR, ROU is represents ROUGE-L, and CID is used for CIDER-D.

| Model | B@1 | B@2 | B@3 | B@4 | MET | ROU | CID |
|---|---|---|---|---|---|---|---|
| SA [5] | 0.700 | 0.490 | 0.322 | 0.242 | 0.238 | - | - |
| ME [26] | 0.731 | 0.559 | 0.429 | 0.299 | 0.246 | 0.529 | 1.001 |
| ATT [12] | 0.699 | 0.527 | 0.399 | 0.299 | 0.232 | - | - |
| SC [15] | 0.719 | 0.540 | 0.400 | 0.297 | 0.239 | - | 0.94 |
| HA[5] | 0.715 | 0.503 | 0.355 | 0.249 | 0.229 | - | - |
| NIC [6] | 0.659 | 0.449 | 0.399 | 0.202 | - | - | - |
| LRCN [41] | 0.690 | 0.514 | 0.379 | 0.270 | 0.230 | 0.500 | 0.830 |
| LSTM-A5 | 0.729 | 0.559 | 0.429 | 0.325 | 0.253 | 0.539 | 1.002 |
| **Proposed CNN+GRU** | **0.751** | **0.578** | **0.439** | **0.335** | **0.259** | **0.545** | **1.035** |

- Test results on COCO's online test server

Table 2: Performance comparisons on online COCO test server (C40). MS Captivator is a photo caption model suggested by Fang et al. [27].

| Model | B@1 | B@2 | B@3 | B@4 | MET | ROU | CID |
|---|---|---|---|---|---|---|---|
| SA [5] | 0.721 | 0.494 | 0.333 | 0.251 | 0.242 | 0.511 | 0.98 |
| ME [26] | 0.743 | 0.561 | 0.432 | 0.293 | 0.251 | 0.534 | 1.013 |
| ATT [12] | 0.691 | 0.532 | 0.393 | 0.287 | 0.242 | - | - |
| SC [15] | 0.729 | 0.544 | 0.412 | 0.281 | 0.241 | 0.511 | 0.92 |
| HA[5] | 0.725 | 0.512 | 0.358 | 0.253 | 0.231 | - | - |
| NIC [6] | 0.669 | 0.456 | 0.382 | 0.218 | 0.232 | - | - |
| LRCN [41] | 0.698 | 0.521 | 0.382 | 0.281 | 0.241 | 0.512 | 0.812 |
| LSTM-A5 | 0.731 | 0.562 | 0.432 | 0.331 | 0.258 | 0.542 | 1.000 |
| **Proposed** | **0.742** | **0.583** | **0.441** | **0.339** | **0.263** | **0.556** | **1.012** |

Table 3: Performance comparisons on online COCO test server (C40). MS Captivator is a photo caption model suggested by Fang et al. [27].

| Model | B@1 | B@2 | B@3 | B@4 | MET | ROU | CID |
|---|---|---|---|---|---|---|---|
| SA [5] | 0.898 | 0.800 | 0.605 | 0.505 | 0.347 | 0.686 | 0.940 |
| ME [26] | 0.900 | 0.781 | 0.701 | 0.575 | 0.340 | 0.685 | 0.864 |
| ATT [12] | 0.897 | 0.825 | 0.605 | 0.505 | 0.345 | 0.680 | 0.928 |
| SC [15] | 0.900 | 0.808 | 0.705 | 0.505 | 0.343 | 0.685 | 0.919 |
| HA[5] | 0.899 | 0.804 | 0.622 | 0.515 | 0.345 | 0.682 | 0.940 |
| NIC [6] | 0.901 | 0.809 | 0.711 | 0.510 | 0.337 | 0.680 | 0.952 |
| LRCN [41] | 0.902 | 0.814 | 0.710 | 0.512 | 0.339 | 0.680 | 0.947 |
| LSTM-A5 | 0.903 | 0.816 | 0.702 | 0.602 | 0.338 | 0.686 | 0.964 |
| **Proposed CNN+GRU** | **0.904** | **0.818** | **0.712** | **0.603** | **0.343** | **0.687** | **0.967** |

To further confirm the validity of our model, the COCO's online test server was used to evaluate the performance on the test set. In particular, the captions made by proposed CNN-GRU were uploaded to the server to do the comparisons with the baseline models. The official test images, 5 (c5) reference captions, and 40 (c40) reference captions used in Table 2 and Table 3 are shown. It is clearly seen again on the results that our method outperformed all other state-of-the-art methods. The results of the metreces BLEU@1, BLEU@2, BLEU@3 and BLEU@4 are all better than the NIC, HA, SA, ATT, ME, and other compared methods. Even on the metrics METEOR which is shown as M in Table 2 and Table 3, ROUGE-L which is shown as R and CIDEr-D which is shown as C in the Table 2 and Table 3, which were initially developed as benchmarks for evaluating the accuracy of machine translation, our resulting indexes are still better on the above metrices as compared to NIC, HA, SA, ATT, ME, and other compared methods.

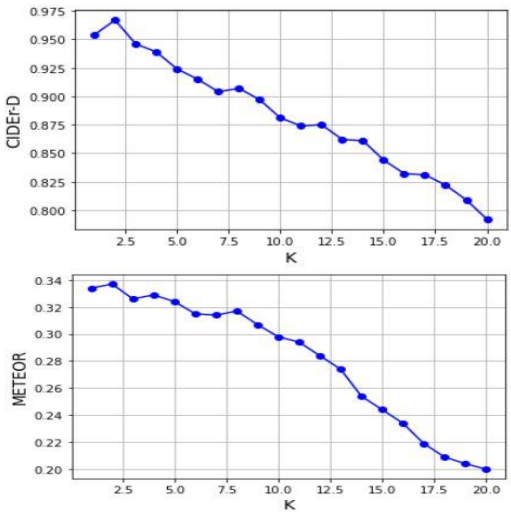

**Fig.3** Model performance with different K.

## 5. CONCLUSION AND FUTURE WORK

In this paper, we have proposed CNN-GRU based hybrid deep learning model with better hyper-parameter tuning to do image captioning. Our CNN-GRU encoder decode framework do caption-to-image reconstruction to handle the semantic context into consideration as well as the time complexity. By taking the hidden states of the decoder into consideration, the input image and its similar semantic representations were reconstructed and reconstruction scores from a semantic reconstructor were used in conjunction with likelihood during model training to assess the quality of the generated caption. As a result, the decoder received improved semantic information, enhancing the caption production process. During model testing, combining the reconstruction score and the log-likelihood was also feasible to choose the most appropriate caption. The suggested model outperforms the state-of-the-art LSTM-A5 model for picture captioning in terms of time complexity and accuracy.

REFERENCES
1. Alahmadi, R., & Hahn, J. (2022). Improve Image Captioning by Estimating the Gazing Patterns from the Caption. In Proceedings of the IEEE/CVF Winter Conference on Applications of Computer Vision (pp. 1025-1034).
2. Wei, J., Li, Z., Zhu, J., & Ma, H. (2022). Enhance understanding and reasoning ability for image captioning. Applied Intelligence, 1-17.
3. Zhao, S., Li, L., & Peng, H. (2022). Aligned visual semantic scene graph for image captioning. Displays, 74, 102210.
4. Singh, A., Krishna Raguru, J., Prasad, G., Chauhan, S., Tiwari, P. K., Zaguia, A., &Ullah, M. A. (2022). Medical Image Captioning Using Optimized Deep Learning Model. Computational Intelligence and Neuroscience, 2022.
5. Bahdanau, D., Cho, K., & Bengio, Y. (2014). Neural machine translation by jointly learning to align and translate. arXiv preprint arXiv:1409.0473.
6. Fukushima, K., & Miyake, S. (1982). Neocognitron: A self-organizing neural network model for a mechanism of visual pattern recognition. In Competition and cooperation in neural nets (pp. 267-285). Springer, Berlin, Heidelberg.
7. He, K., Gkioxari, G., Dollár, P., & Girshick, R. (2017). Mask r-cnn. In Proceedings of the IEEE international conference on computer vision (pp. 2961-2969).
8. Girshick, R. (2015). Fast r-cnn. In Proceedings of the IEEE international conference on computer vision (pp. 1440-1448).
9. Ren, S., He, K., Girshick, R., & Sun, J. (2015). Faster r-cnn: Towards real-time object detection with region proposal networks. Advances in neural information processing systems, 28.
10. Ji, J., Huang, X., Sun, X., Zhou, Y., Luo, G., Cao, L., ...& Ji, R. (2022). Multi-Branch Distance-Sensitive Self-Attention Network for Image Captioning. IEEE Transactions on Multimedia.
11. Wang, X., Wu, J., Chen, J., Li, L., Wang, Y. F., & Wang, W. Y. (2019). Vatex: A large-scale, high-quality multilingual dataset for video-and-language research. In Proceedings of the IEEE/CVF International Conference on Computer Vision (pp. 4581-4591).
12. Cornia, M., Baraldi, L., & Cucchiara, R. (2019). Show, control and tell: A framework for generating controllable and grounded captions. In Proceedings of the IEEE/CVF Conference on Computer Vision and Pattern Recognition (pp. 8307-8316).
13. Herdade, S., Kappeler, A., Boakye, K., & Soares, J. (2019). Image captioning: Transforming objects into words. Advances in Neural Information Processing Systems, 32.
14. Huang, L., Wang, W., Chen, J., & Wei, X. Y. (2019). Attention on attention for image captioning. In Proceedings of the IEEE/CVF International Conference on Computer Vision (pp. 4634-4643).
15. Ke, L., Pei, W., Li, R., Shen, X., & Tai, Y. W. (2019). Reflective decoding network for image captioning. In Proceedings of the IEEE/CVF International Conference on Computer Vision (pp. 8888-8897).
16. Li, G., Zhu, L., Liu, P., & Yang, Y. (2019). Entangled transformer for image captioning. In Proceedings of the IEEE/CVF International Conference on Computer Vision (pp. 8928-8937).
17. Vinyals, O., Toshev, A., Bengio, S., & Erhan, D. (2016). Show and tell: Lessons learned from the 2015 mscoco image captioning challenge. IEEE transactions on pattern analysis and machine intelligence, 39(4), 652-663.
18. Tu, Z., Liu, Y., Shang, L., Liu, X., & Li, H. (2017, February). Neural machine translation with reconstruction. In Thirty-First AAAI Conference on Artificial Intelligence.
19. Mao, J., Xu, W., Yang, Y., Wang, J., Huang, Z., & Yuille, A. (2014). Deep captioning with multimodal recurrent neural networks (m-rnn). arXiv preprint arXiv:1412.6632..
20. Karpathy, A., & Fei-Fei, L. (2015). Deep visual-semantic alignments for generating image descriptions. In Proceedings of the IEEE conference on computer vision and pattern recognition (pp. 3128-3137)..
21. Xu, K., Ba, J., Kiros, R., Cho, K., Courville, A., Salakhudinov, R., ...& Bengio, Y. (2015, June). Show, attend and tell: Neural image caption generation with visual attention. In International conference on machine learning (pp. 2048-2057). PMLR.
22. Vinyals, O., Toshev, A., Bengio, S., & Erhan, D. (2015). Show and tell: A neural image caption generator. In Proceedings of the IEEE conference on computer vision and pattern recognition (pp. 3156-3164).
23. Yang, Z., Yuan, Y., Wu, Y., Cohen, W. W., & Salakhutdinov, R. R. (2016). Review networks for caption generation. Advances in neural information processing systems, 29.
24. Liu, C., Mao, J., Sha, F., & Yuille, A. (2017, February). Attention correctness in neural image captioning. In Proceedings of the AAAI Conference on Artificial Intelligence (Vol. 31, No. 1).
25. Anderson, P., He, X., Buehler, C., Teney, D., Johnson, M., Gould, S., & Zhang, L. (2018). Bottom-up and top-down attention for image captioning and visual question answering. In Proceedings of the IEEE conference on computer vision and pattern recognition (pp. 6077-6086).
26. Jia, X., Gavves, E., Fernando, B., & Tuytelaars, T. (2015). Guiding the long-short term memory model for image caption generation. In Proceedings of the IEEE international conference on computer vision (pp. 2407-2415).
27. Wu, Q., Shen, C., Liu, L., Dick, A., & Van Den Hengel, A. (2016). What value do explicit high level concepts have in vision to language problems? In Proceedings of the IEEE conference on computer vision and pattern recognition (pp. 203-212).
28. Cheng, Y. (2019). Joint Training for Neural Machine Translation. Springer Nature.
29. 19. Gan, Z., Gan, C., He, X., Pu, Y., Tran, K., Gao, J., ...& Deng, L. (2017). Semantic compositional networks for visual captioning. In Proceedings of the IEEE conference on computer vision and pattern recognition (pp. 5630-5639).
30. Yao, T., Pan, Y., Li, Y., Qiu, Z., & Mei, T. (2017). Boosting image captioning with attributes. In Proceedings of the IEEE international conference on computer vision (pp. 4894-4902).
31. Lu, J., Zhou, R., & Luo, S. (2016). Design of Extended 'Digital Communication Theory'Hardware Implementation Platform. In Computer Science and Engineering Technology (CSET2015) & Medical Science and Biological Engineering (MSBE2015) Proceedings of the 2015 International Conference on CSET & MSBE (pp. 289-297).
32. Karpathy, A., & Fei-Fei, L. (2015). Deep visual-semantic alignments for generating image descriptions. In Proceedings of the IEEE conference on computer vision and pattern recognition (pp. 3128-3137)..
33. Xu, K., Ba, J., Kiros, R., Cho, K., Courville, A., Salakhudinov, R., ...& Bengio, Y. (2015, June). Show, attend and tell: Neural image caption generation with visual attention. In International conference on machine learning (pp. 2048-2057). PMLR.
34. Vinyals, O., Toshev, A., Bengio, S., & Erhan, D. (2015). Show and tell: A neural image caption generator. In Proceedings of the IEEE conference on computer vision and pattern recognition (pp. 3156-3164).
35. Papineni, K., Roukos, S., Ward, T., & Zhu, W. J. (2002, July). Bleu: a method for automatic evaluation of machine translation. In Proceedings of the 40th annual meeting of the Association for Computational Linguistics (pp. 311-318).
36. Banerjee, S., & Lavie, A. (2005, June). METEOR: An automatic metric for MT evaluation with improved correlation with human judgments. In Proceedings of the acl workshop on intrinsic and extrinsic evaluation measures for machine translation and/or summarization (pp. 65-72).
37. Lin, C. Y. (2004, July). Rouge: A package for automatic evaluation of summaries. In Text summarization branches out (pp. 74-81).
38. Vedantam, R., Lawrence Zitnick, C., & Parikh, D. (2015). Cider: Consensus-based image description evaluation. In Proceedings of the IEEE conference on computer vision and pattern recognition (pp. 4566-4575).
39. Chen, X., Fang, H., Lin, T. Y., Vedantam, R., Gupta, S., Dollár, P., & Zitnick, C. L. (2015). Microsoft coco captions: Data collection and evaluation server. arXiv preprint arXiv:1504.00325.